\documentclass[conference,10pt]{IEEEtran}
\IEEEoverridecommandlockouts
\usepackage{cite}
\usepackage{amsmath,amssymb,amsfonts}
\usepackage{algorithm}
\usepackage{algpseudocode}
\usepackage{graphicx}
\usepackage{textcomp}
\usepackage{xcolor}
\usepackage{multirow}
\usepackage{subfigure}
\usepackage{subcaption}
\usepackage{adjustbox}
\usepackage{comment}
\usepackage[left=1.62cm,right=1.62cm,top=1.9cm]{geometry}
\def\BibTeX{{\rm B\kern-.05em{\sc i\kern-.025em b}\kern-.08em
  T\kern-.1667em\lower.7ex\hbox{E}\kern-.125emX}}
\begin{document}

\title{Towards Secure and Scalable Energy Theft Detection: A Federated Learning Approach for Resource-Constrained Smart Meters\\
}

\author{\IEEEauthorblockN{1\textsuperscript{st} Diego Labate}
\IEEEauthorblockA{\textit{DIMES} \\
\textit{University of Calabria}\\
Rende, Italy \\
diego.labate@dimes.unical.it}
\and
\IEEEauthorblockN{2\textsuperscript{nd} Dipanwita Thakur}
\IEEEauthorblockA{\textit{DIMES} \\
\textit{University of Calabria}\\
Rende, Italy \\
dipanwita.thakur@unical.it}
\and
\IEEEauthorblockN{3\textsuperscript{rd} Giancarlo Fortino}
\IEEEauthorblockA{\textit{DIMES} \\
\textit{University of Calabria}\\
Rende, Italy \\
giancarlo.fortino@unical.it}

}

\maketitle

\begin{abstract}

Energy theft poses a significant threat to the stability and efficiency of smart grids, leading to substantial economic losses and operational challenges. Traditional centralized machine learning approaches for theft detection require aggregating user data, raising serious concerns about privacy and data security. These issues are further exacerbated in smart meter environments, where devices are often resource-constrained and lack the capacity to run heavy models. In this work, we propose a privacy-preserving federated learning framework for energy theft detection that addresses both privacy and computational constraints. Our approach leverages a lightweight multilayer perceptron (MLP) model, suitable for deployment on low-power smart meters, and integrates basic differential privacy (DP) by injecting Gaussian noise into local model updates before aggregation. This ensures formal privacy guarantees without compromising learning performance. We evaluate our framework on a real-world smart meter dataset under both IID and non-IID data distributions. Experimental results demonstrate that our method achieves competitive accuracy, precision, recall, and AUC scores while maintaining privacy and efficiency. This makes the proposed solution practical and scalable for secure energy theft detection in next-generation smart grid infrastructures. 
\end{abstract}

\begin{IEEEkeywords}
Electricity theft detection, Federated Learning, Smart meters, Differential Privacy
\end{IEEEkeywords}

\section{Introduction}
Energy theft is a pervasive issue across global power grids, resulting in annual losses exceeding $96$ billion worldwide, with developing countries accounting for over 80\% of the total loss \cite{IEA2025global}. As energy systems evolve toward digital smart grids, the integration of smart meters plays a vital role in improving monitoring, billing accuracy, and demand forecasting. However, the increased granularity and connectivity of smart meters have also opened new avenues for malicious activities and privacy violations.
Regions like India, Brazil, and parts of Africa report energy theft rates ranging from 15\% to over 40\% of distributed electricity, largely due to tampering, bypassing meters, or manipulating reported consumption \cite{WorldBank2020losses}. While machine learning (ML) techniques have been explored to detect anomalous consumption patterns indicative of theft, most of these approaches rely on centralized data collection, raising major concerns about user privacy, especially in light of regulations like the EU’s GDPR and India’s Personal Data Protection Bill. Moreover, the heterogeneity and resource constraints of smart meters—often equipped with limited CPU and memory—render many complex ML models impractical for on-device deployment. High-latency and energy costs further discourage frequent data transmission to centralized servers. These challenges necessitate a shift toward decentralized, privacy-aware, and computationally lightweight learning paradigms.
\par
Federated learning (FL) \cite{Mcmahan2017} addresses the problem of data centralization by enabling collaborative training directly on edge devices. Yet, FL alone does not guarantee formal privacy preservation. Differential privacy (DP) \cite{Wei2020} enhances FL by adding mathematically bounded noise to the model updates, significantly reducing the risk of sensitive data leakage—even in the event of server compromise. This work is thus driven by the dual need to detect energy theft accurately while preserving data privacy and accommodating hardware limitations. By integrating basic DP mechanisms into a lightweight MLP model within an FL framework, we aim to provide a deployable, efficient, and privacy-conscious solution to real-world energy theft detection in smart grids.
\par
These contributions aim to address critical issues related to privacy, data security, and computational constraints in the context of energy theft detection in smart grids listed as follows:
\begin{itemize}
    \item The work proposes a federated learning framework designed specifically for energy theft detection in smart grids, addressing both privacy concerns and computational constraints. A MLP model is introduced as a lightweight solution that is suitable for deployment on low-power smart meters, which are typically resource-constrained. The approach incorporates basic DP by injecting Gaussian noise into local model updates before aggregation, ensuring privacy while maintaining formal privacy guarantees.
    \item The framework is evaluated on a real-world smart meter dataset, testing it under both IID and non-IID data distributions. The experimental results demonstrate that the proposed method achieves competitive performance results, including accuracy, precision, recall, and AUC scores, while maintaining privacy and computational efficiency.
    \item The proposed solution is shown to be practical and scalable for energy theft detection in next-generation smart grid infrastructures, making it suitable for real-world deployment.
\end{itemize}

\section{Related Work}
This section provides a brief overview of recent research, with an emphasis on FL techniques and smart grid energy theft detection systems. 

Several FL frameworks have been proposed for energy theft detection, each with unique strengths and limitations. FedDetect\cite{Wen2022} introduced a comprehensive FL framework for energy theft detection, combining local differential privacy and homomorphic encryption for strong privacy guarantees. It features a secure aggregation protocol without requiring a trusted third party and utilizes a Temporal Convolutional Network (TCN) for accurate detection of time-dependent patterns. However, the limitations include high computational overhead that eliminates the need for a trusted third party and utilizes a Temporal Convolutional Network (TCN) for homomorphic encryption, reduced model accuracy due to data-level differential privacy, and reliance on a semi-trusted multi-entity infrastructure. Additionally, its use of a heavy TCN model makes it less suitable for deployment on resource-constrained smart meters. In \cite{Jithish2023}, the authors presented a FL framework for anomaly detection in smart grids, evaluating seven ML models across multiple datasets in both centralized and FL setups. Using a Raspberry Pi-based testbed, it demonstrates the feasibility of FL on resource-constrained devices by analyzing accuracy and hardware metrics such as CPU, memory, power, and bandwidth. While the approach is comprehensive, it assumes near-IID data across clients, lacks integration of formal differential privacy mechanisms, and does not focus on cost-sensitive learning specific to energy theft. Moreover, performance slightly drops with non-IID distributions, and there's no adaptive method to handle skewed client data. Techniques like SMOTE or noise addition for balancing or privacy are also not explored. FedGrid \cite{Gupta2023} presents a novel FL-based framework designed to enhance smart grid (SG) operations by jointly addressing renewable energy generation forecasting and electric load prediction. The framework ensures data privacy by enabling decentralized model training across distributed entities, eliminating the need to share raw data. This approach is particularly valuable for safeguarding sensitive inputs such as user-specific electricity usage and localized weather conditions. By facilitating collaborative learning without compromising privacy, FedGrid offers a scalable and secure solution to key predictive challenges in modern energy systems. However, the major limitation is its reliance on IID data assumptions, which undermines its applicability to real-world smart grid environments where data is typically non-IID due to diverse user behaviors, regional weather variations, and infrastructure heterogeneity.

In \cite{Nadeem2024}, the authors proposed a privacy-preserving FL method (FL-CNN) for energy theft detection, combining a cost-sensitive loss, SMOTE-based class balancing, and a bias-aware CNN to improve accuracy and reduce financial loss. While effective, the method faces scalability issues, assumes IID data, lacks formal differential privacy, and does not address temporal patterns or resource constraints on edge devices. The authors in \cite{Wen2025} introduced a FL framework for electricity theft detection using a CNN-LSTM model, homomorphic encryption, and prototype-based class imbalance handling. It achieves strong performance and privacy guarantees but suffers from high computational overhead, reliance on trusted third parties, and limited suitability for resource-constrained smart meters. 
\begin{table*}[ht]
\centering
\caption{Comparison of Related Federated Learning Methods for Energy Theft Detection}
\resizebox{0.95\textwidth}{!}{%
\begin{tabular}{|p{3.1cm}|p{3.5cm}|p{3.5cm}|p{3.5cm}|p{3.5cm}|p{3.5cm}|p{3.5cm}|}
\hline
\textbf{Aspect} & \textbf{FedDetect \cite{Wen2022}} & \textbf{FL-CNN \cite{Nadeem2024}} & \textbf{HeteroFL \cite{Wen2025}} & \textbf{\cite{Jithish2023}} & \textbf{FedGrid \cite{Gupta2023}} & \textbf{Our Proposed Method} \\
\hline
Model Type & Temporal Convolutional Network (TCN) & CNN-B (Convolutional with bias correction) & Hybrid CNN-LSTM & Multiple models (LogReg, FFNN, 1D-CNN, Autoencoder, RNN, LSTM, GRU) & DSS-LSTM (Deep State Space LSTM) for forecasting & Lightweight MLP (Multi-layer Perceptron) \\
\hline
Privacy Mechanism & Local Differential Privacy (K-RR) + Homomorphic Encryption + CL* Signatures & No formal privacy mechanism; relies on data locality & CKKS Homomorphic Encryption + Shamir Secret Sharing & SSL/TLS encryption on parameter updates & Federated Learning (privacy by design) only; no DP or HE used & Basic Differential Privacy (Gaussian noise on model updates) \\
\hline
Secure Aggregation & Cryptographic protocol using ECC and bilinear pairing & Standard FedAvg in Flower framework & Multi-party secure protocol with secret sharing & SSL-secured FedAvg (no cryptographic aggregation) & Standard FedAvg; secure transport assumed but not formalized & Standard federated averaging with DP noise \\
\hline
Class Imbalance Handling & Not addressed explicitly & SMOTE + weighted loss & Attentional prototype learning + class-weighted loss & SMOTE-NC for mixed feature types & Not addressed & Not addressed explicitly \\
\hline
Data Distribution Support & IID only & IID only (after SMOTE) & Non-IID via clustering (k-means) & Both IID and non-IID (via random sampling) & IID only (not tested with non-IID) & Both IID and non-IID via sampling \\
\hline
Hardware Suitability & Mid-range edge servers or DTSs with cryptographic capabilities & Edge devices (Raspberry Pi-level) & Requires moderate compute (smart substations) & Raspberry Pi (smart meter prototype) & Tested on NVIDIA Jetson Nano; targets mid-tier edge hardware & Optimized for resource-constrained smart meters \\
\hline
Infrastructure Requirements & Control Center + Data Center + DTS; certificate-based trust & Requires FL server using Flower & Assumes regulatory authority for clustering/label weights & Light FL server + Raspberry Pi clients (low-cost setup) & FL server + multiple utility datasets; centralized coordination & Fully decentralized; no trusted infrastructure needed \\
\hline
Communication Overhead & High (encrypted model updates + certificates) & Moderate (FedAvg updates) & High (encrypted updates + secure masks) & Low to moderate (model updates over SSL) & Moderate (FedAvg, no compression or quantization) & Low (DP-noised plaintext updates) \\
\hline
Deployment Complexity & High (encryption, secure key exchanges, certificates) & Moderate (standard FL framework + CNN tuning) & High (encryption + attention + trusted authority) & Moderate (TensorFlow + Flower + SSL setup) & Moderate (custom LSTM + evaluation on three real datasets) & Low (simple PyTorch-based FL) \\
\hline
Energy Efficiency & Not optimized for energy efficiency & Measured CPU, memory, bandwidth, and power use & Not optimized for edge efficiency & Measured (memory, CPU, power usage optimized for Raspberry Pi) & Not explicitly optimized for energy use & Designed for low-power smart meters \\
\hline
\end{tabular}
}
\label{tab:fl_comparison}
\end{table*}

Table \ref{tab:fl_comparison} provides a comprehensive comparison of recent FL methods proposed for energy theft and anomaly detection in smart grids. However, we consider FedDetect \cite{Wen2022} and FL-CNN \cite{Nadeem2024} as the baseline to compare with our proposed methods, as both used the same data set for the evaluation.

\section{Methodology}

\begin{figure}
    \centering
    \includegraphics[width=0.8\linewidth]{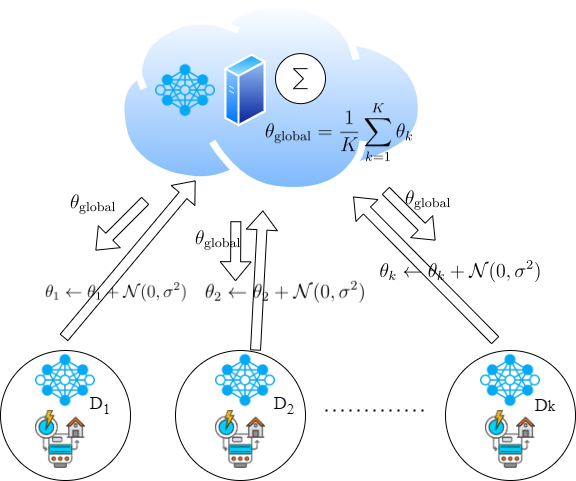}
    \caption{FL Framework}
    \label{fig:fl}
\end{figure}

This study applies an FL framework as shown in Figure \ref{fig:fl} for binary classification in a privacy-preserving setting. The key components of the methodology are outlined below.

\subsection{Data Preprocessing}

Let the dataset be denoted by:
\[
\mathcal{D} = \{(x_i, y_i)\}_{i=1}^N, \quad x_i \in \mathbb{R}^d, \quad y_i \in \{0, 1\}
\]
Features are standardized using z-score normalization:
\[
x_i' = \frac{x_i - \mu}{\sigma}, \quad \forall i \in \{1, \dots, N\}
\]
The dataset is partitioned into training and testing sets using an 80:20 split:
\[
\mathcal{D}_{\text{train}}, \mathcal{D}_{\text{test}} = \text{split}(\mathcal{D})
\]

\subsection{Client Partitioning}

Assume $K$ clients ($K = 2, 3 \text{and} 5$ in our implementation). The training set is distributed among the clients as follows:

\begin{itemize}
  \item \textbf{IID Partition:}
  \[
  \mathcal{D}_{\text{train}}^{(k)} \sim \text{Uniform Partition}(\mathcal{D}_{\text{train}})
  \]
  \item \textbf{Non-IID Partition:} Data is grouped by class labels and distributed so that each client receives samples from a limited subset of the label distribution.
\end{itemize}
\subsection{Model Architecture}

MLP is used as the local and global model. The forward pass is defined as:
\begin{align*}
f_\theta(x) &= \text{Softmax}\Big(W_3 \cdot \text{ReLU}\Big( 
    W_2 \cdot \text{ReLU}\big( W_1 x + b_1 \big)\\ + b_2
\Big)\Big)
\end{align*}
where:
\begin{align*}
  W_1 &\in \mathbb{R}^{128 \times d}, & W_2 &\in \mathbb{R}^{64 \times 128}, & W_3 &\in \mathbb{R}^{2 \times 64} \\
  b_1 &\in \mathbb{R}^{128}, & b_2 &\in \mathbb{R}^{64}, & b_3 &\in \mathbb{R}^{2}
\end{align*}

\subsection{Local Training}

Each client trains its model locally using Cross-Entropy Loss:
\[
\mathcal{L}_{\text{CE}}(\theta) = -\sum_{i=1}^{B} y_i \log \hat{y}_i,
\] where $B$ is the batch size.

The optimizer used is Stochastic Gradient Descent (SGD) with a decaying learning rate:
\[
\eta_r = \eta_0 \cdot \gamma^r, \quad \text{where } r \text{ is the communication round}
\]

\subsection{Federated Averaging}

After local training, model parameters from each client $k$ are aggregated at the server via federated averaging:
\[
\theta_{\text{global}} = \frac{1}{K} \sum_{k=1}^{K} \theta_k
\]

To simulate privacy-preserving noise injection, Gaussian noise is optionally added:
\[
\theta_k \leftarrow \theta_k + \mathcal{N}(0, \sigma^2), \quad \text{where } \sigma \text{ is the noise standard deviation}
\]
\subsection{Communication Cost and Bandwidth Reduction}

In FL, communication overhead is a significant factor compared to centralized training. The communication cost primarily arises from transmitting model parameters between the server and clients over multiple rounds.

Let $P$ denote the total number of model parameters, $B$ be the size in bytes of each parameter (typically $B = 4$ for 32-bit floats), $K$ be the number of clients, and $R$ be the total number of communication rounds. The total communication cost is given by:
\[
\text{Cost}_{\text{FL}} = 2 \times R \times K \times P \times B
\]
The factor of 2 accounts for both uploading local models and downloading the global model in each round.

In contrast, centralized training requires each client to transmit its entire local dataset $\mathcal{D}^{(k)}$ to a central server. The cost of centralized data upload is:
\[
\text{Cost}_{\text{centralized}} = \sum_{k=1}^{K} |\mathcal{D}^{(k)}| \times d \times B
\]
where $|\mathcal{D}^{(k)}|$ is the number of samples held by client $k$ and $d$ is the feature dimensionality.

The bandwidth reduction achieved by FL compared to centralized training is computed as:
\[
\text{Bandwidth Reduction} = 1 - \frac{\text{Cost}_{\text{FL}}}{\text{Cost}_{\text{centralized}}}
\]

This metric quantifies the efficiency of the proposed method in minimizing data transmission, making it suitable for privacy-sensitive and resource-constrained environments.

\subsection{Evaluation Metrics}

The global model is evaluated on the test dataset using the following metrics:

\begin{itemize}
  \item \textbf{Accuracy}:
  \[
  \text{Accuracy} = \frac{TP + TN}{TP + TN + FP + FN}
  \]
  \item \textbf{Precision}:
  \[
  \text{Precision} = \frac{TP}{TP + FP}
  \]
  \item \textbf{Recall}:
  \[
  \text{Recall} = \frac{TP}{TP + FN}
  \]
\item \textbf{F1-weighted}:
    \[
\text{F1}_{\text{weighted}} = \frac{\sum_{i=1}^{C} s_i \cdot \text{F1}_i}{\sum_{i=1}^{C} s_i}
\]

\noindent
\text{where:} \\
\quad $C$ \text{ is the number of classes,} \\
\quad $s_i$ \text{ is the support (number of true instances) of class } i, \\
\quad $\text{F1}_i$ \text{ is the F1-score for class } i.
  
  \item \textbf{AUC} (Area Under the ROC Curve):
  \[
  \text{AUC} = \int_{0}^{1} \text{TPR}(FPR^{-1}(x)) \, dx
  \]
\end{itemize}

\subsection{Visualization and Logging}

The training process is logged over multiple communication rounds, and performance is visualized using accuracy/loss plots and ROC curves. All results are saved to disk for post-analysis.

\subsection{Federated Learning Algorithm}

\begin{algorithm}[H]
\caption{Federated Learning with MLP}\label{algo:algo1}
\begin{algorithmic}[1]
\State \textbf{Input:} Training dataset $\mathcal{D}_{\text{train}}$, number of clients $K$, number of rounds $R$, local epochs $E$, learning rate $\eta$, noise standard deviation $\sigma$
\State Partition $\mathcal{D}_{\text{train}}$ into $K$ subsets: $\mathcal{D}^{(1)}, \dots, \mathcal{D}^{(K)}$
\State Initialize global model $\theta^{(0)}$
\For{$r = 1$ to $R$}
  \For{each client $k \in \{1, \dots, K\}$ \textbf{in parallel}}
    \State $\theta^{(r,k)} \leftarrow \theta^{(r-1)}$  \Comment{Initialize local model}
    \For{each local epoch $e = 1$ to $E$}
      \State Update $\theta^{(r,k)}$ using SGD on $\mathcal{D}^{(k)}$
    \EndFor
    \State Optionally add noise: $\theta^{(r,k)} \leftarrow \theta^{(r,k)} + \mathcal{N}(0, \sigma^2)$
  \EndFor
  \State Aggregate: $\theta^{(r)} \leftarrow \frac{1}{K} \sum_{k=1}^{K} \theta^{(r,k)}$
\EndFor
\State \textbf{Return:} Final global model $\theta^{(R)}$
\end{algorithmic}
\end{algorithm}

The presented algorithm \ref{algo:algo1} outlines the used FL framework with a MLP model. It begins by dividing the training dataset into $K$ subsets, each assigned to a different client. A global model is initialized and iteratively improved over $R$ communication rounds. In each round, every client starts with the current global model and locally updates it for $E$ epochs using SGD on their private data. Optionally, Gaussian noise with standard deviation 
$\sigma$ can be added to enhance privacy. After all clients finish their local training, their models are averaged to update the global model. This process continues until the final global model is obtained after $R$ rounds.
\section{Experimental Results}
\begin{table}[ht]
    \centering
    \resizebox{\linewidth}{!}{  
    \begin{tabular}{|c|c|c|c|}
    \hline
    \textbf{Clients} & \textbf{Distribution} & \textbf{Clients} & \textbf{Samples}\\
    \hline
   \multirow{ 4}{*}{2}  & \multirow{ 2}{*}{IID}   &Client 0  & 12696 (Label 0: 11643, Label 1: 1053)\\
      & & Client 1 & 12696 (Label 0: 11671, Label 1: 1025)\\
     
      \cline{2-4}
     &  \multirow{ 2}{*}{non-IID}  & Client 0  & 12696 (Label 0: 11657, Label 1: 1039)\\
     &  & Client 1 & 12696 (Label 0: 11657, Label 1: 1039)\\
       
    \hline
   \multirow{ 6}{*}{3}  & \multirow{ 3}{*}{IID}   &Client 0  & 8464 (Label 0: 7752, Label 1: 712)\\
      & & Client 1 & 8464 (Label 0: 7785, Label 1: 679)\\
      &  & Client 2 & 8464 (Label 0: 7777, Label 1: 687)\\
      \cline{2-4}
     &  \multirow{ 3}{*}{non-IID}  & Client 0  & 8465 (Label 0: 7772, Label 1: 693)\\
     &  & Client 1 & 8464 (Label 0: 7771, Label 1: 693)\\
     &   & Client 2 & 8463 (Label 0: 7771, Label 1: 692)\\
        \hline
        
   \multirow{ 10}{*}{5}  & \multirow{ 5}{*}{IID}   &Client 0  & 5079 (Label 0: 4656, Label 1: 423)\\
      & & Client 1 & 5079 (Label 0: 4658, Label 1: 421)\\
      &  & Client 2 & 5078 (Label 0: 4667, Label 1: 411)\\
      & & Client 3 & 5078 (Label 0: 4660, Label 1: 418) \\
     & & Client 4 & 5078 (Label 0: 4673, Label 1: 405)\\
      \cline{2-4}
     &  \multirow{ 5}{*}{non-IID}  & Client 0  & 5079 (Label 0: 4663, Label 1: 416)\\
     &  & Client 1 & 5079 (Label 0: 4663, Label 1: 416)\\
     &   & Client 2 & 5079 (Label 0: 4663, Label 1: 416)\\
     & & Client 3 & 5078 (Label 0: 4663, Label 1: 415)\\
     & & Client 4 & 5077 (Label 0: 4662, Label 1: 415)\\
        \hline
    \end{tabular}
    }
    \caption{Dataset distributions among clients (devices)}
    \label{tab:datadistribution}
\end{table}
We conduct the experiment on a realistic electricity consumption dataset, which was released by  State Grid Corporation of China (SGCC) \cite{Zheng2018}. Specifically, this dataset includes information on 42,372 power users' electricity use for a period of 1035 days (January 1, 2014 to October 31, 2016). In addition to the publicly available dataset, the SGCC made it clear that 3615 electricity thieves are included in it. Since power thieves account for about 9\% of all customers, it is clear that there is a significant problem with electricity theft in China.  When assessing the effectiveness of the suggested plan and any comparable programs, we use the provided electricity thieves as the ground truth. It is important to note that there are some inaccurate and missing numbers in the dataset. The data set is divided into 3 different clients (devices) for both the IID and non-IID. For non-IID distribution, label based partitioning is used, which results in client optimizing their local models with skewed label distributions, making global aggregation more challenging due to conflicting gradient updates. In this work, we consider full participation of the clients. Table \ref{tab:datadistribution} shows the data distributions among the clients for our experiment.
\par
\begin{figure*}[!ht]
\subfigure[Accuracy-loss with IID]{\includegraphics[width=0.23\textwidth]{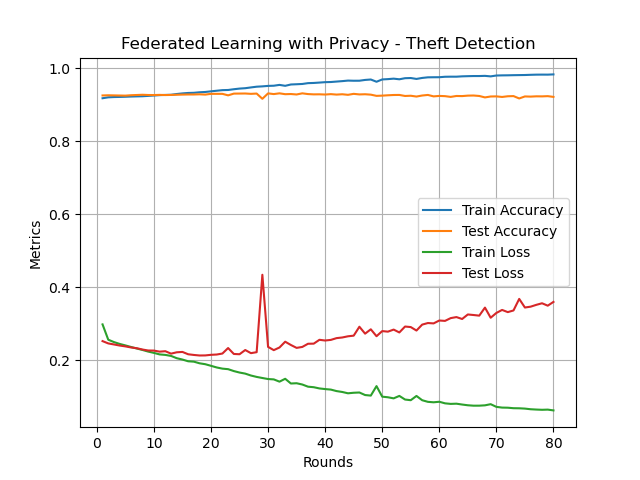}}\quad
\subfigure[ROC with IID]{\includegraphics[width=0.23\textwidth]{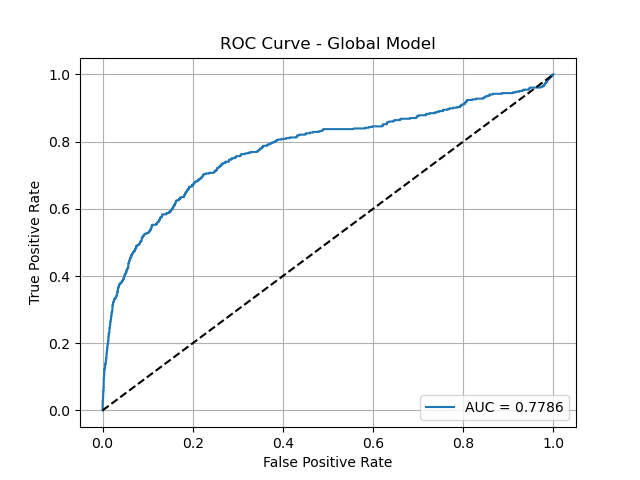}}\quad
\subfigure[Accuracy-loss with Non-IID]{\includegraphics[width=0.23\textwidth]{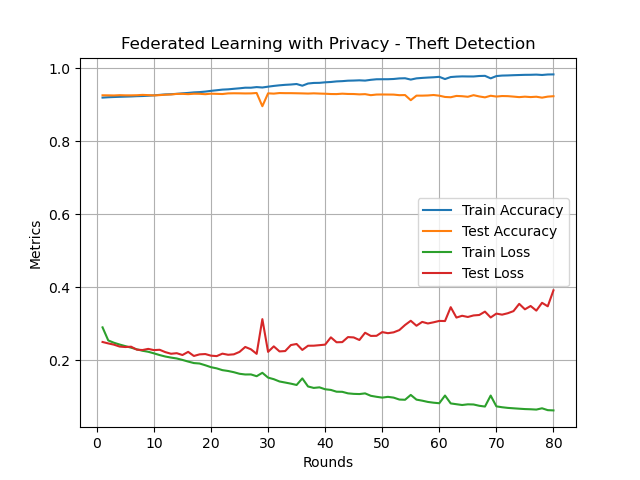}}\quad
\subfigure[ROC with non-IID]{\includegraphics[width=0.23\textwidth]{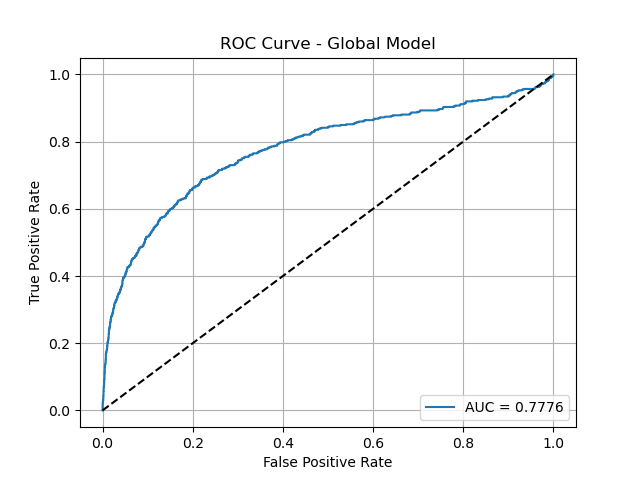}}
\caption{Performance of 2 clients for 3 epochs and 80 rounds}
\label{fig:clients2_round80_epochs3}
\end{figure*}%

\begin{figure*}[!ht]
\subfigure[Accuracy-loss with IID]{\includegraphics[width=0.23\textwidth]{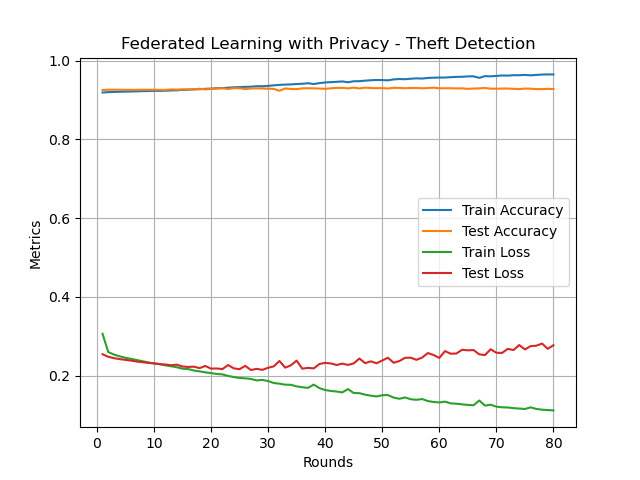}}\quad
\subfigure[ROC with IID]{\includegraphics[width=0.23\textwidth]{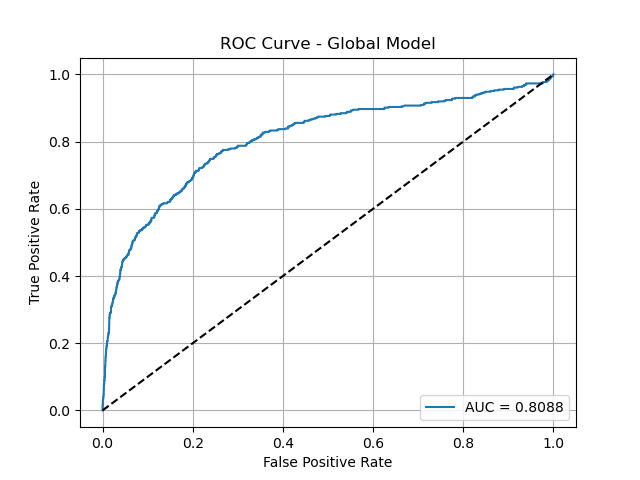}}\quad
\subfigure[Accuracy-loss with Non-IID]{\includegraphics[width=0.23\textwidth]{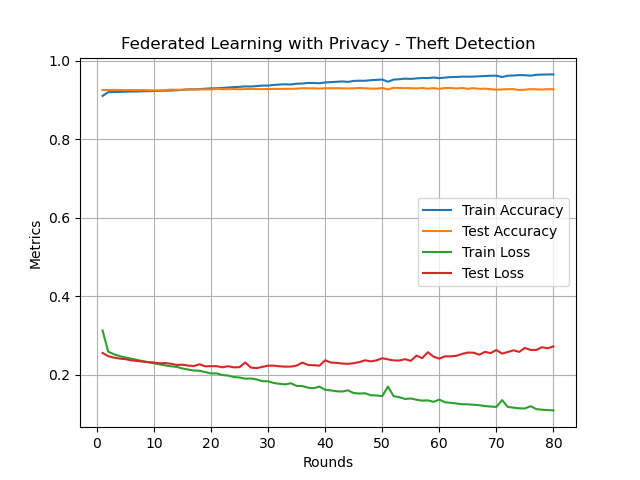}}\quad
\subfigure[ROC with non-IID]{\includegraphics[width=0.23\textwidth]{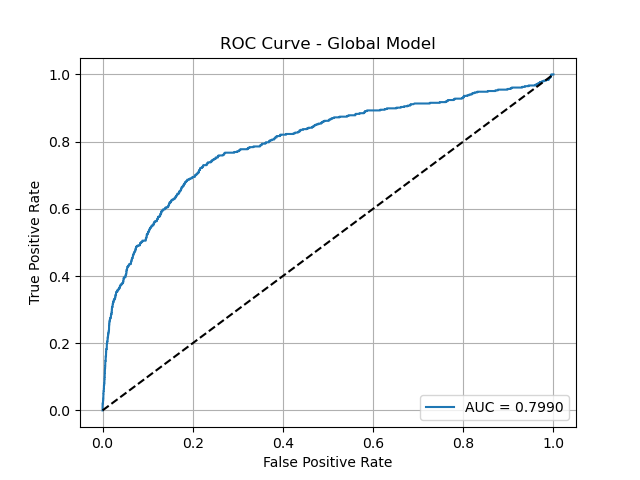}}
\caption{Performance of 3 clients for 3 epochs and 80 rounds}
\label{fig:clients3_round80_epochs3}
\end{figure*}%

\begin{figure*}[!ht]
\subfigure[Accuracy-loss with IID]{\includegraphics[width=0.23\textwidth]{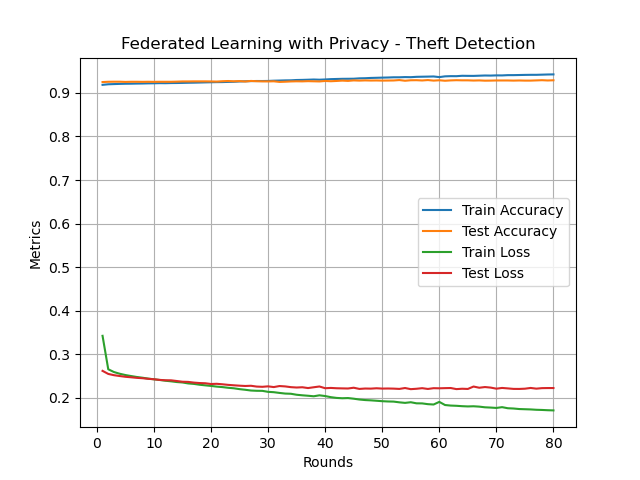}}\quad
\subfigure[ROC with IID]{\includegraphics[width=0.23\textwidth]{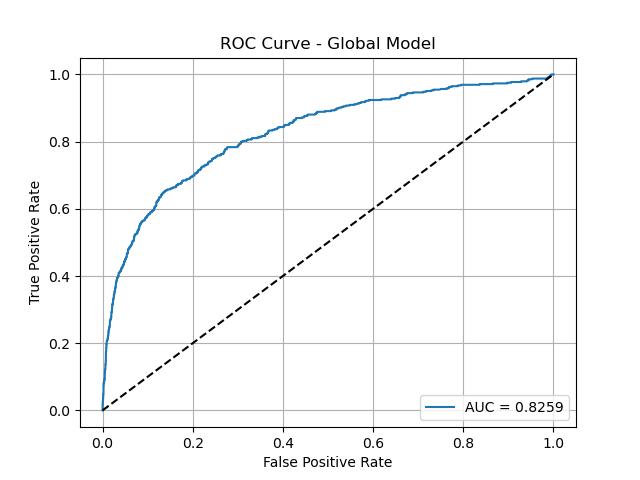}}\quad
\subfigure[Accuracy-loss with Non-IID]{\includegraphics[width=0.23\textwidth]{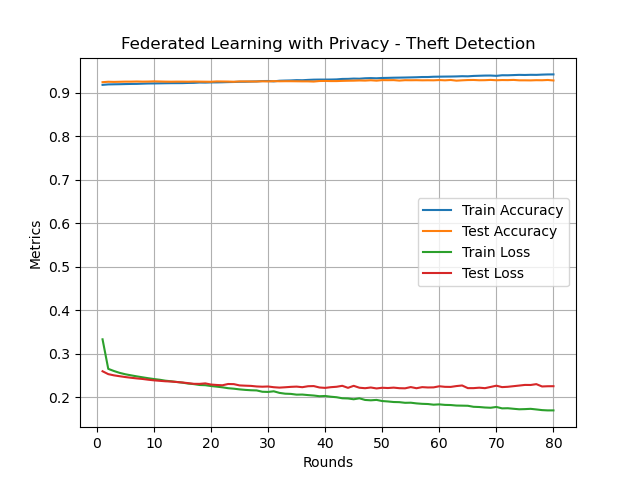}}\quad
\subfigure[ROC with non-IID]{\includegraphics[width=0.23\textwidth]{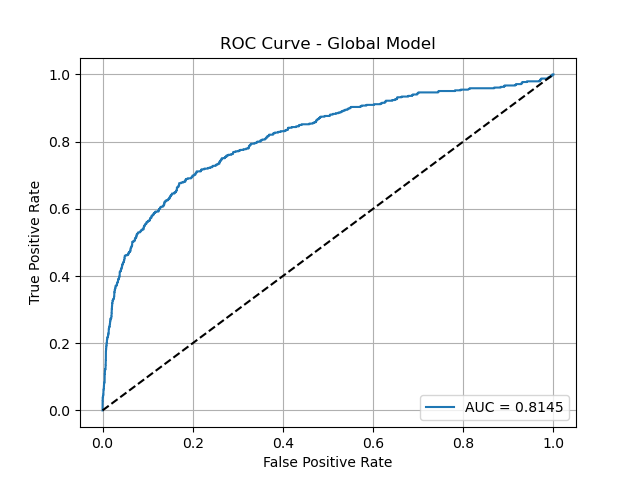}}
\caption{Performance of 5 clients for 3 epochs and 80 rounds}
\label{fig:clients5_round80_epochs3}
\end{figure*}

The classification model employed in this study is a MLP, serving as both the local client model and the global aggregated model. It consists of three fully connected layers: an input layer projecting the $d$-dimensional input to 128 hidden units, followed by a second layer reducing the dimensionality to 64 units, and finally an output layer producing logits for the two target classes. Non-linear activation is introduced via the ReLU function after each hidden layer, and a Softmax function is applied to the final output for probability estimation. The model parameters include weight matrices $W_1, W_2, W_3$ and bias vectors $b_1, b_2, b_3$, all of which are updated during local training and aggregated during the federated rounds. This architecture balances simplicity and expressive capacity, making it well-suited for distributed learning across heterogeneous devices. The models are implemented using PyTorch (v.2.0), and the experiments are conducted on a computer equipped with an Intel(R) Core(TM) i5-10210U CPU @ 1.60GHz, Windows-11-10.0.26100-SP0, 11.851 GB RAM, and 8 CPU count.

Table \ref{tab:datadistribution} shows the label-based distribution values in run-time. In this work, no data balancing approach is applied; keep in mind that resource-constrained edge devices are not powerful enough to compute much in real-time. However, the performance as shown in the Figures \ref{fig:clients2_round80_epochs3}, \ref{fig:clients3_round80_epochs3}, and \ref{fig:clients5_round80_epochs3} with an imbalance dataset for 2, 3, and 5 clients is competitive enough with state-of-the-art (SOTA) methods. The results are shown for 80 rounds and 3 epochs. In this experiment, SGD optimization algorithm is used with a decaying learning rate. The initial learning rate is 0.01. Empirical evaluations confirm the effectiveness and robustness of the method, as the global model consistently improves in both accuracy and loss across communication rounds. The ability to scale the system by simply increasing the number of participating clients demonstrates the framework’s adaptability to larger smart grid networks. Collectively, these characteristics make the proposed approach not only practical and privacy-aware but also scalable and efficient for next-generation energy monitoring systems.

Moreover, the performance of the proposed method is shown using different performance metrics such as loss, accuracy, precision, recall, and AUC-ROC. In addition, we also calculate the reduction in communication and bandwidth (BW) in MegaBytes (MB) as shown in Table \ref{tab:SOTA}. The performance of our proposed method outperforms the SOTA methods.
\begin{table*}[!ht]
    \centering
    \resizebox{\linewidth}{!}{  
    {\bf
    \begin{tabular}{c|c|c|c|c|c|c|c|c|c}
    \hline
        Reference & Clients & Loss & Accuracy & Precision & Recall & F1-Score & AUC-ROC & Communication Cost (MB) & BW Reduction (MB) \\
        \cline{1-9}
        Centralized MLP & - & 0.2439 & 91.93 & 65.62 & 11.62 & 19.75 & 81.50 & -&100.06\\
        \hline
       
       \multirow{ 2}{*}{FLDetect\cite{Wen2022}} & 2 & - & 91.98 & 91.00 & 90.98 & 40.18 & - & -& 57.0\\
       & 5 & - & 92.58 & 91.03 & 92.58 & 36.07 & - & -& 10.32 \\
      \hline
       \multirow{ 2}{*}{FL-CNN\cite{Nadeem2024}} & 2 & - & 87.75 & 91.90 & 87.75 & 49.29 & - & -&69.8\\
       & 5 & - & 80.53 & 90.34 & 80.53 & 37.11 & - & -& 23.12 \\
       
    \hline    
     \multirow{ 3}{*}{Our Proposed Method} & 2 (IID) & 0.0962 & 97.69& 88.89 & 82.05 & 97.65 & 98.12& 122.07& 37.93\\
     & 2 (non-IID) & 0.0648 & 98.22& 99.01 & 78.92 & 98.12 & 98.33 & 122.07& 37.93\\
       & 3 (IID) & 0.1112 & 96.49 &94.66  & 60.59 &96.14  & 95.94 & 183.11& 28.97 \\
       & 3 (non-IID) & 0.1120 & 96.42 & 96.43 & 58.42 & 96.01 & 95.62 & 183.11& 28.97 \\
        & 5 (IID) &0.1750 &  94.08  & 85.96 &33.01  & 92.84 &89.42 &305.18& 21.49 \\
         & 5 (non-IID) & 0.1748 & 94.17 & 87.34 & 33.88 &92.98  & 89.70 & 305.18& 21.49 \\
    \hline    
    \end{tabular}
    }
    }
    \caption{Comparison with SOTA methods}
    \label{tab:SOTA}
 
\end{table*}
To benchmark our approach, we compare our proposed method against some of the recently proposed methods on the same dataset as shown in \ref{tab:SOTA}. The FL communication cost and BW reduction are more due to the increased number of rounds. We perform 80 rounds to analyse the performance of the proposed method. Though we achieve higher test accuracy and lower cost within 10 rounds. 
Almost all the methods did not consider the non-IID distributions and the resource-constrained limitations of the edge devices. For example, compared to the FL-CNN model in \cite{Nadeem2024}, our approach differs in its simplicity, modularity, and focus on evaluating the core aspects of FL, such as model generalization, data distribution effects, and communication efficiency. While FL-CNN integrates energy-aware cost functions and advanced bias correction through SMOTE and custom loss weighting, our model emphasizes practical deployment with minimal computational and communication overhead, making it more accessible to resource-constrained environments. However, our method does not yet incorporate temporal patterns, cost-based penalties, or advanced differential privacy frameworks, which are addressed in part by the FL-CNN architecture. Similarly, compared to FedDetect \cite{Wen2022}, which employs a heavy TCN model with complex privacy mechanisms such as local differential privacy and homomorphic encryption, our proposed method uses a lightweight MLP optimized for edge devices. While FedDetect assumes moderate resources and a multi-layered security infrastructure, our approach targets resource-constrained smart meters with simple FL and optional Gaussian noise for differential privacy. Additionally, our method explicitly supports non-IID data distribution and minimizes communication and computation overhead, making it more suitable for practical, decentralized deployment scenarios. On the other hand, the HeteroFL framework proposed by Wen et al. \cite{Wen2025} utilizes a CNN-LSTM model with homomorphic encryption and attentional prototype learning for electricity theft detection. While their method achieves high detection accuracy, it incurs significant computational and communication overhead, making it less suitable for deployment on edge devices. In contrast, our proposed method, based on a lightweight MLP architecture with differential privacy, achieves competitive accuracy while significantly reducing resource consumption. Additionally, our framework demonstrates better scalability and faster convergence under both IID and non-IID settings, confirming its practical advantage for real-world smart meter applications.

It is pertinent to note that, as shown in Table \ref{tab:SOTA}, the model performs better with non-IID data than with IID data, despite theory suggesting that IID data should be easier to learn from. However, any oversampling or rebalancing is applied in this dataset. In non-IID, each client might train primarily on one class (e.g., label 0 or 1). This forces the global model to learn class-specific patterns more robustly from each client. Hence, the global model may develop sharper class boundaries, improving precision and recall. Moreover, the simple MLP model, DP, which adds noise and non-IID data with a class-specific focus, regularizes the learning and prevents overfitting on overly smooth, well-mixed IID data. Hence, it gives better test generalization from diverse, focused local models. Another reason for the better performance of the non-IID data can be dilution of the minority class. If the theft class (label 1) is rare, in an IID split, each client might only get a few minority-class examples, and local models won’t learn meaningful theft patterns, then the performance decreases. In contrast, non-IID clients might get more concentrated theft samples (even accidentally), which improves learning. Moreover, the energy reduction is the same for IID and non-IID data distributions as the formula does not depend on the data distribution (IID or non-IID). It only cares about model size, number of rounds, and number of clients, which remain constant in both setups.

\section{Conclusion and Future Work}
In this study, we proposed a privacy-preserving federated learning framework for detecting energy theft in smart meter networks. Our approach leverages lightweight MLPs to accommodate resource-constrained devices, while employing differential privacy mechanisms to ensure client data confidentiality during collaborative training. Experimental results on real-world smart meter data demonstrate that our framework achieves promising performance in terms of accuracy, precision, recall, and AUC, despite the challenges posed by non-IID data distributions and strict privacy constraints. The integration of differential privacy introduces noise into model updates, slightly impacting model performance, but offering strong privacy guarantees. Overall, our method presents a viable solution for secure and scalable energy theft detection in smart grids, particularly in environments where centralized data collection is infeasible or raises privacy concerns. Moreover, the architecture of the proposed framework supports horizontal scalability, enabling effective integration of an increasing number of clients without degrading model performance or communication efficiency. This inherent adaptability ensures the framework remains robust and performant across large-scale smart grid infrastructures. As a result, the solution is not only privacy-preserving and computationally efficient but also architecturally suited for widespread deployment in next-generation energy monitoring and theft detection systems.

Future work will focus on several key directions. First, we aim to explore adaptive privacy budgets and personalized noise injection to better balance privacy and performance per client. Second, we plan to extend our framework to support asynchronous federated learning and incorporate more realistic scenarios with client dropout, varying energy constraints, and communication delays. Finally, we are investigating the integration of explainable AI techniques to enhance the interpretability of the theft detection decisions, which is crucial for stakeholder trust and transparency in smart grid systems.

\section*{Acknowledgment}

This work contributes to the basic research activities of the PNRR project FAIR -  Future AI Research (PE00000013), Spoke 9 - Green-aware AI, under the NRRP MUR program funded by the NextGenerationEU.

\bibliographystyle{IEEEtran}
\bibliography{references}

@misc{IEA2025global,
  author       = {{International Energy Agency}},
  title        = {Global Energy Review 2025},
  year         = {2025},
  howpublished = {\url{https://www.iea.org/reports/global-energy-review-2025}},
  note         = {Accessed: 2025-04-08}
}

@misc{WorldBank2020losses,
  author       = {{World Bank}},
  title        = {Reducing Technical and Non-Technical Losses in Power Networks},
  year         = {2020},
  howpublished = {\url{https://openknowledge.worldbank.org/handle/10986/34524}},
  note         = {Accessed: 2025-04-08}
}

@ARTICLE{Zheng2018,
  author={Zheng, Zibin and Yang, Yatao and Niu, Xiangdong and Dai, Hong-Ning and Zhou, Yuren},
  journal={IEEE Transactions on Industrial Informatics}, 
  title={Wide and Deep Convolutional Neural Networks for Electricity-Theft Detection to Secure Smart Grids}, 
  year={2018},
  volume={14},
  number={4},
  pages={1606-1615},
  }

@INPROCEEDINGS{Nadeem2024,
  author={Nadeem, Zunaira and Jaber, Mona},
  booktitle={2024 IEEE 99th Vehicular Technology Conference (VTC2024-Spring)}, 
  title={Privacy Preserving Energy-Aware Federated Learning Based Method for Energy Theft Detection}, 
  year={2024},
  volume={},
  number={},
  pages={1-7},
  }

@ARTICLE{Wen2022,
  author={Wen, Mi and Xie, Rong and Lu, Kejie and Wang, Liangliang and Zhang, Kai},
  journal={IEEE Internet of Things Journal}, 
  title={FedDetect: A Novel Privacy-Preserving Federated Learning Framework for Energy Theft Detection in Smart Grid}, 
  year={2022},
  volume={9},
  number={8},
  pages={6069-6080}
  }

@article{Wen2025,
title = {A privacy-preserving heterogeneous federated learning framework with class imbalance learning for electricity theft detection},
author = {Hanguan Wen and Xiufeng Liu and Bo Lei and Ming Yang and Xu Cheng and Zhe Chen},
journal = {Applied Energy},
volume = {378},
pages = {124789},
year = {2025},
issn = {0306-2619}
}

@ARTICLE{Jithish2023,
  author={Jithish, J. and Alangot, Bithin and Mahalingam, Nagarajan and Yeo, Kiat Seng},
  journal={IEEE Access}, 
  title={Distributed Anomaly Detection in Smart Grids: A Federated Learning-Based Approach}, 
  year={2023},
  volume={11},
  number={},
  pages={7157-7179},
  }

@inproceedings{Mcmahan2017,
  title={Communication-efficient learning of deep networks from decentralized data},
  author={McMahan, Brendan and Moore, Eider and Ramage, Daniel and Hampson, Seth and y Arcas, Blaise Aguera},
  booktitle={Artificial intelligence and statistics},
  pages={1273--1282},
  year={2017},
  organization={PMLR}
}

@ARTICLE{Wei2020,
  author={Wei, Kang and Li, Jun and Ding, Ming and Ma, Chuan and Yang, Howard H. and Farokhi, Farhad and Jin, Shi and Quek, Tony Q. S. and Vincent Poor, H.},
  journal={IEEE Transactions on Information Forensics and Security}, 
  title={Federated Learning With Differential Privacy: Algorithms and Performance Analysis}, 
  year={2020},
  volume={15},
  number={},
  pages={3454-3469},
  }

@Article{Gupta2023,
AUTHOR = {Gupta, Harshit and Agarwal, Piyush and Gupta, Kartik and Baliarsingh, Suhana and Vyas, O. P. and Puliafito, Antonio},
TITLE = {FedGrid: A Secure Framework with Federated Learning for Energy Optimization in the Smart Grid},
JOURNAL = {Energies},
VOLUME = {16},
YEAR = {2023},
NUMBER = {24},
ARTICLE-NUMBER = {8097},
URL = {https://www.mdpi.com/1996-1073/16/24/8097},
ISSN = {1996-1073},
}

\end{document}